\documentclass{article}

\usepackage[margin=1in]{geometry}

\usepackage{natbib}
\usepackage[utf8]{inputenc} %
\usepackage[T1]{fontenc}    %
\usepackage{hyperref}       %
\usepackage{url}            %
\usepackage{booktabs}       %
\usepackage{amsfonts,amsthm}       %
\usepackage{nicefrac}       %
\usepackage{microtype,xcolor}      %
\usepackage{amsfonts,amsmath,amssymb}      
\usepackage{bbm}
\usepackage{algorithm}
\usepackage[noend]{algpseudocode}
\usepackage{graphicx, subcaption}

\newtheorem{prop}{Proposition}

\newcommand{\MC}[1]{{\color{red} MC: #1}}
\newcommand{\JP}[1]{{\color{purple} JP: #1}}

\newcommand {\OMIT}[1]{}               %

\newcommand {\br}[1]{\left(#1\right)}
\newcommand {\sqb}[1]{\left[#1\right]}
\newcommand {\cbr}[1]{\left\{#1 \right\}}

\newcommand{\RR}{\mathbb{R}}    %
\newcommand{\NN}{\mathbb{N}}    %

\newcommand{\Mcal}{\mathcal{M}}

\newcommand{\Xcal}{\mathcal{X}} %

\newcommand{\argmax}{\mathop{\mathrm{argmax}\,}}
\newcommand{\trace}{\mathop{\text{trace}\,}}

\newcommand\E{\mathbb E}

\definecolor{darkgreen}{rgb}{0,0.5,0.0}
\renewcommand\phi\varphi

\newcommand\One{\mathbf{1}}
\newcommand{\eqdef}{\ensuremath{\stackrel{\mbox{\upshape\tiny def.}}{=}}}
\DeclareMathOperator{\KL}{KL}
\renewcommand*{\thefootnote}{\arabic{footnote}}

\title{Deep multi-class learning from label proportions}

\makeatletter
\renewcommand*\@fnsymbol[1]{\the#1}
\makeatother
\author{
  Gabriel Dulac-Arnold\,\footnote{Google Research, Brain Team}$^{\,\,\,,*}$\\
  \texttt{dulacarnold@google.com}
  \and
  Neil Zeghidour\,$^{\text{1},*}$\\
  \texttt{neilz@google.com}
  \and
  Marco Cuturi\,$^\text{1,}$\footnote{CREST ENSAE, F-91120 Palaiseau, France}\\
  \texttt{cuturi@google.com}
  \and
  Lucas Beyer\,$^\text{1}$\\
  \texttt{lbeyer@google.com}
  \and
  Jean-Philippe Vert\,$^\text{1,}$\footnote{MINES ParisTech, PSL University, CBIO - Centre for Computational Biology, F-75006 Paris, France}\\  
  \texttt{jpvert@google.com}
}

\date{}
\begin{document}

\maketitle
\let\thefootnote\relax\footnotetext{$^*$Equal contribution}
\setcounter{footnote}{1}

\begin{abstract}
    We propose a learning algorithm capable of learning from label proportions instead of direct data labels.  In this scenario, our data are arranged into various bags of a certain size, and only the proportions of each label within a given bag are known. This is a common situation in cases where per-data labeling is lengthy, but a more general label is easily accessible. Several approaches have been proposed to learn in this setting with linear models in the multiclass setting, or with nonlinear models in the binary classification setting. Here we investigate the more general nonlinear multiclass setting, and compare two differentiable loss functions to train end-to-end deep neural networks from bags with label proportions. We illustrate the relevance of our methods on an image classification benchmark, and demonstrate the possibility to learn accurate image classifiers from bags of images.
\end{abstract}

\section{Introduction}

The standard setting of supervised classification in machine learning assumes that we have access to a training set of samples and to their labels; our goal is then to estimate a classifier able to predict the label of new samples. In many real-world situations, however, collecting training sets of labeled examples is not possible, and alternative learning scenarios must be considered. We focus in this paper on a particular setting where one has access to \emph{bags} of examples, and where for each bag only the proportions of the labels in the bag are available; the task is still to learn a classifier to predict the label of individual samples. This setting, which following~\citet{Yu20013SVM} we refer to as \emph{learning from label proportions} (LLP), is relevant in many situations where labeling of individual samples is time-consuming, difficult, or just not possible, while side-channel information can be used to reconstruct the proportions of label within a given bag. For example, \citet{Musicant2007Supervised} explain how LLP is a natural setting to analyze single particle mass spectrometry data, while~\citet{Quadrianto2009Estimating} discuss applications in e-commerce, politics or spam filtering. LLP is particularly relevant in situations where labels are only provided at an aggregated level for privacy-preserving reasons, as in medical databases, fraud detection or election results, as reviewed by~\citet{Patrini2014Almost}, or in computer vision applications for visual attribute modeling~\citep{Chen2014Object,Yu2014Modeling}, event detection in videos~\citep{Lai2014Video} or classification of synthetic aperture radar (SAR) images \citep{Ding2017Learning}. More generally, with recent advances in deep learning, it is clear that collecting very large amounts of labeled data is a good recipe for success in many applied fields, but for a long tail of applications (e.g., medical images) only experts in the field are knowledgeable enough to annotate data. For such applications, collecting aggregated annotations, such as a rough estimate of the proportion of cancer cells in an image, is more realistic than asking an expert to label individually thousands or millions of individual cells.

The LLP setting has attracted increasing attention in the machine learning community recently, and a number of methods have been proposed. \citet{Musicant2007Supervised} introduced the LLP problem and proposed that standard algorithms for supervised classification, such as support vector machines (SVM), $k$-nearest neighbors (kNN) and multilayer perceptrons can be adapted to the LLP setting by a slight modification of the objective functions of these algorithms. In parallel, \citet{deFreitas2005Learning} also introduced the LLP scenario and proposed a MCMC-based hierarchical Bayesian model which generates labels consistent with the proportions, which however does not scale well to large datasets; \citet{Hernandez-Gonzalez2013Learning} also proposed an MCMC-based Bayesian approach, which suffers from the same limitations. \citet{Quadrianto2008Estimating,Quadrianto2009Estimating} proposed the \emph{mean map} model (MeanMap), which is based on strong modelling assumptions including the fact that the data follow an exponential model, and that the class-conditional distribution of data is independent of the bags. \citet{Fan2014Learning} proposed a variant of MeanMap, while \citet{Patrini2014Almost} extended MeanMap to more general objective functions; these family of methods are however tailored to the situation where the number of bags is of the order of the number of classes, and where the model learned is linear in some fixed feature space. \citet{Chen2009Kernel} and \citet{Stolpe2011Learning} used $k$-means clustering to identify a clustering of the data compatible with the label proportions, under the assumption that data in each class form clusters that can be captured by $k$-means. \citet{Rueping2010SVM} proposed a method called \emph{inverse calibration} (InvCal) that adapts SVM to the LLP setting, which however is restricted to linear models in some feature space. \citet{Yu20013SVM} proposed another extension of SVM, called $\propto$SVM, which iteratively fits an instance-level classifier and estimates the labels of individual samples; the method is computationally efficient in the binary classification setting, and was later extended to other binary classifiers~\citep{Wang2015Linear,Li2015Alter,Chen2017Learning,Qi2017LEarning,Shi2019Learning,Shi2018Learning}. This family of methods, however, can not easily be extended to non-binary multiclass classification problems, since they rely on some sorting operations specific to the binary classification case. \citet{Kotzias2015From} proposed to optimize an instance-level classifier by minimizing a bag-level loss (how much the distribution of predictions differs from the known distribution), penalized by a regularization term that enforces similar instances to share similar classes. However, it is limited to learning a linear model on a fixed vector representation of the data. \citet{Bortsova2018Deep} propose to train a deep neural network for binary classification by penalizing in the loss function how much the proportion of samples of each class in a bag differs from the bag proportion, and \citet{Ardehaly2017Co} follow a similar strategy to train a convolution neural network in a multi-class setting.

Apart from~\citet{Ardehaly2017Co}, and to the best of our knowledge, there is no previous work on LLP when we want to learn a non-binary classification model using deep learning (DL), a setting of immense practical interest given the remarkable performance of DL on numerous tasks in computer vision or natural language processing, for example. Most existing approaches either rely heavily on specific, non DL-based models such as linear models in some feature space \citep{Quadrianto2009Estimating,Patrini2014Almost,Kotzias2015From} or $k$-means clustering \citep{Chen2009Kernel,Stolpe2011Learning}, or on the setting of binary classification to derive efficient algorithms \citep{Yu20013SVM,Shi2018Learning}.

In this work, we propose to use DL-based models for LLP in the multi-class classification setting, by considering two differentiable loss functions that can be used to optimize any standard DL model for individual instances in the LLP setting. The first loss function directly measures how well the labels predicted by a model for instances in a bag fit the known distribution in a bag. Similar ideas have been proposed in the past, e.g. by \citet{Musicant2007Supervised,Kotzias2015From} in different settings, and recently by \citet{Ardehaly2017Co} in a similar setting. The second loss is new and aims to extend to the multiclass setting the idea underlying $\propto$SVM \citep{Yu20013SVM} and similar approaches~\citep{Wang2015Linear,Li2015Alter,Chen2017Learning,Qi2017LEarning,Shi2019Learning,Shi2018Learning} in the binary classification setting, where an estimate of the individual labels within each bag is jointly optimized with the model during training. While efficient alternative optimization schemes can be derived in the binary classification scheme~\citep{Yu20013SVM}, the direct extension of this idea to the multiclass classification setting results in an untractable combinatorial optimization problem. We overcome this limitation by a convex relaxation and an entropic regularization of the objective function, which results in a differentiable loss function that can be optimized efficiently and backpropagated through our neural architecture thanks to recent advances in computational optimal transport \citep{Cuturi2013Lightspeed,Peyre2019Computational}. We assess empirically the performance of both loss functions on two standard image classification benchmarks (CIFAR10 and CIFAR100) using a modern DL architecture (Resnet18), where we demonstrate that the degradation in performance remains very limited with bags of up to a few tens of images, while it slowly decreases for bags with hundreds of images, highlighting the potential of LLP for state-of-the-art applications. We further demonstrate that both losses lead to overall very similar performance in both experiments, suggesting that jointly estimating individual labels and the model parameters during training may not bring benefits over building a standard bag-level model.

\section{Setting and notations}
$\mathbbm{1(\cdot)}$ denotes the indicator function, taking values $1$ or $0$ depending on whether its argument is true or not. Given a set $S$, we denote by $S^\star = \cup_{i=1}^\infty S^i$ the set of nonempty tuples of elements of $S$. For any integer $n\in\NN$, let $[1,n]=\cbr{1,\ldots,n}$ and $\One_n\in\RR^n$ be the $n$-dimensional vector of ones. Given two vectors $a,b\in\RR^n$, where $b_i\neq 0$ for $i\in[1,n]$, we denote by $a\oslash b \in \RR^n$ the vector with entries $(a\oslash b)_i = a_i / b_i$. For any vector or matrix $M$, we denote by $\log(M)$, $\exp(M)$ or $M^\alpha$ (for $\alpha\in\RR$) the matrices obtained by applying the operation entrywise, e.g., $[\log(M)]_{ij} = \log(M_{ij})$, and by $M^\top$ the transpose of $M$.

We consider a supervised multi-class classification problem, where $\Xcal$ is the space of input data (e.g., $\Xcal=\RR^{32\times 32 \times 3}$ for 3-channel $32\times 32$ images), and $K$ is the number of classes. For any class $i\in[1,K]$ let $e(i) \in \cbr{0,1}^K$ be the one-hot encoded version of $i$, i.e., $e(i)_j=1$ if and only if $e(i)=j$ (for $j\in[1,K]$), and let $E_K = \cbr{e(i)\,:\,i\in[1,K]}$ be the set of one-hot encoded classes, seen as binary vectors in $\RR^K$. Let also $\Delta_K =\cbr{z \in \RR_+^K\,:\, \sum_{i=1}^K z_i = 1}$ be the probability simplex, which is also the convex hull of $E_K$.

Our goal is to learn a classifier $h:\Xcal \rightarrow [1,K]$ to predict one category out of $K$ classes for each sample $x\in\Xcal$. For that purpose, we consider training data in the form of $N$ bags $B_1, \ldots, B_N$, where for each $i\in[1,N]$ the bag $B_i$ is a set of $n_i\geq 1$ labeled samples $B_i = \br{(x_{i,1},y_{i,1}), \ldots, (x_{i,n_i},y_{i,n_i})}$, with $x_{i,j}\in\Xcal$ and $y_{i,j}\in[1,K]$ for each $j\in[1,n_i]$. For $i\in[1,N]$, we further denote by $z_i \in \Delta_K$ the vector of label proportions in the bag $B_i$, i.e.,
$$
\forall (i,j) \in [1,N]\times [1,K]\,,\quad  (z_{i})_j = \frac{1}{n_i} \sum_{k=1}^{n_i}\mathbbm{1}(y_{i,k} = j)\,.
$$
In LLP, we assume that we do not have access to the labels of individual samples within each bag $B_i$, but instead that we have access to the aggregated data $A_i = \br{x_{i,1}, \ldots, x_{i,n_i}, z_i} \in \Xcal^{n_i}\times \Delta_K$, and our goal is to learn $h$ from $A_1, \ldots, A_N$.

Regarding predictive models, we assume that we work with a class of nonlinear functions $\Mcal = \cbr{f_\theta\,:\,\theta\in \Theta\subset \RR^p}$ where for each $\theta\in\Theta$, $f_\theta:\Xcal\rightarrow\Delta_K$, and we assume that for any $x\in\Xcal$, $\theta \mapsto f_\theta(x)$ is differentiable almost everywhere. In practice, $\Mcal$ can for example represent a deep neural network where $\theta$ represents the weights of the network. A classifier $h$ is readily obtained from a predictor $f_\theta$ by taking $h(x) \in \argmax_{i\in[1,K]} f_{\theta}(x)_i$.

Given $i\in[1,K]$ and any $\theta\in\Theta$, we denote by $F_i(\theta)\in\br{\Delta_K}^{n_i}$ the $K\times n_i$ matrix $\sqb{f_\theta(x_{i,1}), \ldots, f_\theta(x_{i, n_i})}$ of predictions of the model $f_\theta$ for the samples in the $i$-th bag.

\section{Method}

We consider empirical risk minimization estimators that estimate a parameter $\hat{\theta}\in\Theta$ by attempting to minimize an empirical risk of the form
\begin{equation}\label{eq:emprisk}
    R_N(\theta) \eqdef \frac{1}{N} \sum_{i=1}^N \ell\br{F_i(\theta), z_i}\,,
\end{equation}
for some loss function $\ell : \br{\Delta_K}^{\star}\times \Delta_K \rightarrow \RR$. The loss function for the $i$-th bag compares the matrix of predictions $F_i(\theta)$ of the model for all individual samples in a bag, to the vector $z_i$ of label proportions of the bag. We now discuss two strategies to define such a loss function.

\subsection{A loss for bag-level predictions}
A first, intuitive approach to create loss functions to define the empirical risk (\ref{eq:emprisk}) is to summarize all predictions for individual samples in a bag in a predicted profile for the bag, and to assess how dissimilar this predicted profile is from the known profile. More formally, let us consider a bag with $n$ samples $A = \br{x_{1}, \ldots, x_{n}, z} \in \Xcal^{n}\times \Delta_K$, and $F(\theta)\in\br{\Delta_K}^{n}$ the matrix of model predictions for samples in the bag, for a given $\theta\in\Theta$. The vector of predicted label proportion is then $(1/n)\sum_{i=1}^{n} f_\theta(x_{i}) = F(\theta) \One_{n} / n$, and given any divergence $d_1 : \Delta_K^2 \rightarrow \RR$ to compare distributions, we can define the loss functions:
$$
\ell^{\text{prop}}\br{F(\theta), z} = d_1\br{z , F(\theta) \One_{n} / n} \,.
$$
Such losses have appeared previously in the literature as building blocks of LLP models, typically by taking an $L_1$ or $L_2$ distance as divergence \citep{Musicant2007Supervised,Kotzias2015From}. Since we focus on multiclass classification, we take the standard cross-entropy (or Kullback-Leibler divergence) loss in our experiments, which we refer to as the \emph{KL loss} in the rest of the paper; this loss was also considered by \citet{Ardehaly2017Co} in a similar setting.
$$
\ell^{\text{KL}}\br{F(\theta), z} = - \sum_{i=1}^K z_i \log \sqb{F(\theta) \One_{n} / n}_i \,.
$$

\subsection{A combinatorial loss based on individual predictions}
While a loss for bag-level predictions is theoretically sufficient to learn a bag-level classifier~\citep{Yu2014On}, several authors have observed that improved performance can result from guessing the individual labels of samples in each bag \citep{Yu20013SVM,Wang2015Linear,Li2015Alter,Chen2017Learning,Qi2017LEarning,Shi2019Learning,Shi2018Learning}. More precisely, considering again a bag $A = \br{x_{1}, \ldots, x_{n}, z} \in \Xcal^{n}\times \Delta_K$ with $n$ samples and $F(\theta)\in\br{\Delta_K}^{n}$ the matrix of  predictions for a model $f(\theta)$, let us introduce a new vector $t \in [1,K]^{n}$ to represent our ``guesses'' of the individual sample labels in the bag. Good guesses should have two properties. On the one hand, they should be coherent with our model predictions, in the sense that $f_\theta(x_{i})_{t_{i}} = F(\theta)_{t_i,i}$ should be large for $i\in[1,n]$. On the other hand, good guesses should be coherent with the bag label proportions, in the sense that $d_K\br{(1/n)\sum_{i=1}^n e(t_{i}),z}$ should be small, for some divergence $d_K$ on the simplex. Both goals can be combined in a single objective function as follows:
\begin{equation}\label{eq:combloss}
    \ell^{\text{comb}}\br{F(\theta), z} = \min_{t \in [1,K]^{n}} \cbr{-\frac{\alpha}{n} \sum_{i=1}^{n} \log f_\theta(x_{i})_{t_{i}} + (1-\alpha) d_K\br{\frac{1}{n}\sum_{i=1}^n e(t_{i}),z} }\,,
\end{equation}
where $0\leq\alpha\leq 1$ controls the balance between both terms. Note that the first term corresponds to the standard mean negative log-likelihood of the model.

To simplify notations, we can rewrite (\ref{eq:combloss}) in terms of the $K\times n$ matrix of one-hot encoded version of the labels $U \in (E_K)^{n}$. For that purpose, let us introduce the $K\times n$ matrix $C(\theta) = - \log F(\theta)$ where the $\log$ is understood entrywise, i.e., with entries $C(\theta)_{ij} = -\log f_\theta(x_{j})_i$ for $(i,j)\in[1,K]\times [1,n]$. We can then rewrite the loss (\ref{eq:combloss}) more compactly as follows:
\begin{equation}\label{eq:combloss2}
    \ell^{\text{comb}}\br{F(\theta), z} = \min_{U \in (E_K)^{n}} \cbr{\frac{\alpha}{n} \trace\br{C(\theta)^\top U}  + (1-\alpha) d_K\br{U \One_n / n, z}}\,.
\end{equation}

Unfortunately, (\ref{eq:combloss}-\ref{eq:combloss2}) is in general a combinatorial problem which can not be solved by a computationally efficient algorithm. A notable exception exits in the binary classification case ($K=2$), where (\ref{eq:combloss}) can be solved efficiently by first sorting the $n$ values $f_{\theta}(x_{i})_2 - f_{\theta}(x_{i})_1$, for $i\in[1,n]$, then assigning samples to class $2$ from the top to the bottom of this list, and setting the threshold between class $2$ and class $1$ in that list when the minimum of (\ref{eq:combloss}) is reached. This operation has a $O(n \log (n))$ computational complexity because of the need to sort the values, while the search for the threshold is just a linear-time operation, and was exploited by a number of methods for binary LLP classification~\citep{Yu20013SVM,Wang2015Linear,Chen2017Learning,Qi2017LEarning,Shi2019Learning,Shi2018Learning}. However, in the more general multi-class case ($K>2$), this strategy does not work.

\subsection{Convex relaxation in the transportation polytope}
We propose to render (\ref{eq:combloss2}) computationally tractable by relaxing the discrete constraint on $U\in (E_K)^{n}$, considering instead a set of soft-labels  $U\in \br{\Delta_K}^{n}$. This relaxation leads to the following loss:
\begin{equation}\label{eq:relaxloss}
    \ell^{\text{relax}}\br{F(\theta), z} = \min_{U \in \br{\Delta_K}^{n}} \cbr{\frac{\alpha}{n} \trace\br{C(\theta)^\top U}  + (1-\alpha) d_K\br{ U \One_{n}/n, z} }\,.
\end{equation}
Problem (\ref{eq:relaxloss}) is now a convex optimization problem in $U$, which can in principle be solved by a variety of methods to compute $\ell^{\text{relax}}\br{F(\theta), z}$. In addition, the gradient of $\ell^{\text{relax}}\br{F(\theta), z}$ with respect to the matrix $C(\theta)$ is then equal to $\alpha U/n$, allowing to back-propagate the gradient of the risk (\ref{eq:emprisk}) to optimize $\theta$ once we solve (\ref{eq:relaxloss}) for each bag. Since (\ref{eq:relaxloss}) must be solved for each bag at each epoch of the optimization over $\theta$, it is crucial for practical purpose to derive fast solvers for (\ref{eq:relaxloss}).

In the particular case where we want to enforce that the label proportions in the bag are \emph{exactly} respected by the estimated soft-labels $U$, we can take the indicator divergence $d_K(u,v)=0$ if $u=v$, $+\infty$ otherwise. In that case, (\ref{eq:relaxloss}) becomes a linear program (taking $\alpha=1$):
\begin{equation}\label{eq:relaxloss2}
    \ell^{\text{relax-LP}}\br{F(\theta), z} = \min_{U \in \br{\Delta_K}^{n}, U \One_{n} = nz} \frac{1}{n}\trace\br{C(\theta)^\top U} \,,
\end{equation}
where we recognize an optimal transport (OT) problem over $U$ with marginals $\One_n$ and $nz$, and cost matrix $C(\theta)/n$. The solution to this linear program can therefore be found efficiently by OT solvers. Interestingly, the relaxation is tight in this case, in the sense that any solution $U^*$ of (\ref{eq:relaxloss2}) satisfies $U^* \in \br{E_K}^{n}$ and is therefore also a solution of (\ref{eq:combloss2}); in other words, $\ell^{\text{relax}} = \ell^{\text{comb}}$ in that case.

In the more general case where $d_K$ is a divergence on the simplex, then (\ref{eq:relaxloss}) becomes an unbalanced OT problem \citep{Peyre2019Computational}. The relaxation is generally not tight in that case, and fast LP solvers for OT problems can not be used anymore to efficiently solve (\ref{eq:relaxloss}). We therefore resort now to an additional modification of the loss leading to computational benefits.

\subsection{Unbalanced transport with entropic regularization}
A practical approach to solve equation~\eqref{eq:relaxloss} is to add an entropic penalty to the objective function and to consider for the divergence $d_K$ a Kullback-Leibler divergence. In which case the problem becomes
\begin{equation}\label{eq:unbalancedLOSS}
    \ell^{\text{relax-ent}}\br{F(\theta), z} = \min_{U \in \br{\Delta_K}^{n}} \cbr{\frac{\alpha}{n} \left(\trace\br{C(\theta)^\top U} -\varepsilon H(U)\right)   + (1-\alpha) \KL\br{ U \One_{n}/n\,|\,z}}\,,
\end{equation}
where 
$$H(U)\eqdef -\trace(U^T \left(\log(U)-1\right)),\quad \KL(a,b)\eqdef \sum_i a_i \log(a_i/b_i)-a_i+b_i,$$ 
are respectively the Shannon entropy of $U$ and the generalized Kullback-Leibler divergence between two nonnegative vectors. The benefit of using this formulation is that the solution of the optimization problem has a particularly simple form:
\begin{prop}\label{prop:fixedpoints}
The solution $U^*$ of (\ref{eq:unbalancedLOSS}) satisfies $U^* = \textrm{diag}(a) K \textrm{diag}(b)$ for $a\in\RR^K$ and $b\in\RR^n$ which satisfy
$$
\begin{cases}
a &= \br{nz \oslash Kb}^\tau \,,\\
b &= \One_n \oslash K^\top a\,,
\end{cases}
$$
where $K=F^{1/\epsilon}$ and $\tau=\br{1+\alpha\epsilon/(1-\alpha)}^{-1}$.
\end{prop}
To solve (\ref{eq:unbalancedLOSS}), we therefore propose to use a generalized version of the Sinkhorn algorithm for unbalanced OT~\citep{FrognerNIPS,2015-chizat-unbalanced}, detailed in Algorithm~\ref{algo:sinkhorn}. The procedure is an iterative algorithm through which one can backpropagate gradients, which converges to the solution of (\ref{eq:unbalancedLOSS}) when the number of iterations increases. Note that we approximate this solution  using a finite number of Sinkhorn iterations~\citep{adams2011ranking,hashimoto2016learning,2016-bonneel-barycoord,Flamary2018}, which can be themselves backpropagated at little overhead cost. For numerical stability purpose, our implementation of Algorithm~\ref{algo:sinkhorn} is done in the log-domain, as explained for example in~\citep[p.77]{Peyre2019Computational}. The complexity of each iteration is $O(Kn)$.
\begin{algorithm}[t]
\caption{Compute a differentiable approximation to $\ell^{\text{relax-ent}}$}\label{algo:sinkhorn}
\hspace*{\algorithmicindent} \textbf{Input} $F \in (\Delta_K)^n$, $z\in\Delta_K$, $0\leq\alpha\leq 1$, $\epsilon>0$, $n_{iter}\in\NN$\\
\hspace*{\algorithmicindent} \textbf{Output} Differentiable approximation to $\ell^{\text{relax-ent}}(F,z)$
\begin{algorithmic}[1]
\State $K \gets F^{1/\epsilon}$
\State $\tau=\br{1+\alpha\epsilon/(1-\alpha)}^{-1}$
\State $b \gets \One_n$
\For{$i=1$ to $n_{iter}$}
\State $a \gets \br{nz \oslash Kb}^{\tau}$
\State $b \gets \One_n \oslash K^\top a$
\EndFor
\State $U \gets \textrm{diag}(a) K \textrm{diag}(b)$
\State \textbf{return} $\frac{\alpha}{n} \left(-\trace\br{\log(F)^\top U} -\varepsilon H(U)\right)   + (1-\alpha) \KL\br{ U \One_{n}/n\,|\,z}$
\end{algorithmic}
\end{algorithm}

In the rest of the paper, we refer to $\ell^{\text{relax-ent}}$ simply as the Relax-OT (ROT) loss. Interestingly, for bags of size $1$,  the ROT loss boils down to the standard cross-entropy loss, i.e., learning with the ROT loss boils down to standard learning when we have access to labels of individual instances:
\begin{prop}\label{prop:singleton}
For a bag $(x,z)\in\Xcal\times\E_K$ of size $n=1$,
$$
\ell^{\text{relax-ent}}\br{F(\theta), z} = \alpha \ell^{\text{KL}}\br{F(\theta), z} = -\alpha \sum_{i=1}^K z_i \log f_\theta(x)_i \,.
$$
\end{prop}

\OMIT{
We define first a kernel matrix $K(\theta)\eqdef \exp(-C(\theta)/\varepsilon) = F(\theta)^{1/\epsilon}$ (where the $1/\epsilon$ exponent is applied entrywise to the matrix $F(\theta)$):
$$
b_0=\One_{n}, \text{ for } l\geq1, a_{l+1}= \left(nz\oslash (K b_l)\right)^{\tau}, \quad b_{l+1}=\One_{n}\oslash (K^\top a_{l+1})\,,
$$ 
where $\tau=(1+\varepsilon n\alpha/(1-\alpha))^{-1}$ to recover in the limit $l\rightarrow \infty$ that the optimal (partial) transport plan $U$ is therefore $\textrm{diag}(a_l)K\textrm{diag}(b_l)$. Note that we approximate this solution  using a finite number of Sinkhorn iterations~\citep{adams2011ranking,hashimoto2016learning,2016-bonneel-barycoord,Flamary2018}, which can be themselves backpropagated at little overhead cost.\MC{important point: are you interested in differentiating the entire $\ell_i^\text{relax}$ or just the value of the trace. If that's the trace then it's equally easy to differentiate in fact} \JP{We're interested in the loss}
} %

\section{Experiments}\label{sec:exp}
We evaluate and compare the KL and ROT losses for LLP on the standard image classification task of the CIFAR10 and CIFAR100 datasets \citep{cifar}. CIFAR10 contains 60,000 RGB images of size 32x32 from 10 classes, with 6,000 examples per class. We use the standard split to train our models on 50,000 images and test them on the remaining 10,000. CIFAR100 also has 60,000 RGB images, with a similar split, but with 100 classes and 600 images per class.

\subsection{Experimental setup}
{\bf Training data.} We adapt the standard CIFAR10 and CIFAR100 datasets to the LLP setting by deriving a bag-level supervision from labeled instances. For a given bag size $n$, we sample $n$ examples without replacement from the training set, in an uniform way. This bag is then stored as a training instance, as we do not resample bags during training. We compute as many bags as possible from the original training set, and label them with the vector of label proportions within the bag. In our experiments, we train our systems with bag sizes $n \in \{1, 2, 4, 8, 16, 32, 64, 128, 256, 512, 1024\}$.

{\bf Architecture and training.} Our main architecture is a Residual Network \citep{resnet} with 18 layers (Resnet-18). We do not use biases except for the last fully connected layer. All our models are trained with stochastic gradient descent (SGD) \citep{sgd}, with a momentum of $0.9$. We experiment with learning rates in $\{0.001, 0.003, 0.01, 0.03, 0.1, 0.3\}$. The loss is averaged over a mini-batch, which can contain several bags. All our models are trained for $400$ epochs and we divide the initial learning rate by $10$ mid-training. The weights of the network are learned with a weight decay factor of $0.005$. To avoid overfitting, we furthermore perform a standard data augmentation procedure: when a batch is fed to the network, each of its images is randomly shifted by one pixel, and randomly flipped on the vertical axis with a probability $0.5$.

{\bf Hyperparameters of the ROT loss.} The ROT loss has several hyperparameters, in particular the weight $\alpha$ which controls the trade-off between the coherence of the latent vector with the model's prediction, and its similarity to the real proportion within the bag. We experiment with $\alpha \in \{0.1, 0.3, 0.5, 0.7, 0.9\}$. To compute the loss, we perform 75 iterations of the Sinkhorn algorithm. The weight $\epsilon$ of the entropy term is fixed to $1$.

{\bf Baseline method.} To assess the relevance of using losses for bags, we add as a baseline method a neural network trained on individual images, where we assign to each image the label distribution of the bag it belongs to, using the KL loss for each sample. In other words, compared to the bag-level KL loss $\ell^{\text{KL}}$, we compute the mean of the cross-entropy over samples in a bag, instead of the cross-entropy of the mean prediction. 
\begin{figure}[t]
    \centering
    \includegraphics[width=1\textwidth, trim={4cm 0 0 0},clip]{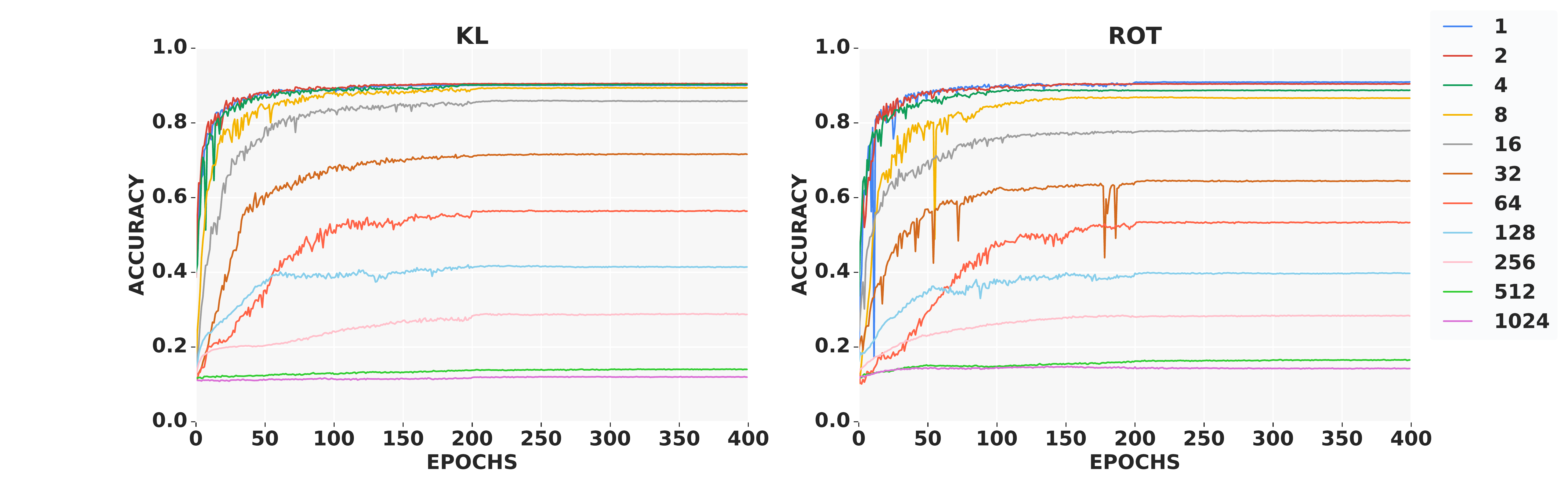}
    \caption{Evolution of the accuracy on the test set of CIFAR10 for various bag sizes, for models trained with a KL-divergence (left, KL) or with optimal transport (right, ROT).}
    \label{fig:ot_and_kl_vs_bag_size}
\end{figure}

\subsection{Results}
{\bf Performance over bag sizes} Figure \ref{fig:ot_and_kl_vs_bag_size} shows the accuracy for both the KL and the ROT loss functions on the test set of CIFAR10. As expected, the instance-level accuracy of our models reaches high accuracy (>90\%) for bags of size $1$, corresponding to the standard image classification setting, and degrades as the bag size grows. Interestingly, we observe that for bag sizes up to $8\sim 16$, the performance is only mildly or not affected when compared to the fully supervised topline. This suggests that for small bag sizes, the LLP setting can be efficiently addressed by both the KL and ROT loss functions. This also indicates that if an artificial bag-level labeling was designed to preserve instance-level privacy during training, large bags should be used for the anonymization not to be decyphered too easily. For bigger bags, the accuracy of our models degrades slowly and steadily, losing about $15\%$ in accuracy each time the bag size doubles, and reaching an accuracy close to chance level ($10\%$ accuracy on CIFAR10) for bag sizes above $1024$ . This can be expected, as with growing size $n$, the distribution of labels inside every bag converges to the same value $\One_{n}/n$, which cannot be disambiguated by a classifier.

\begin{figure}
    \centering
\begin{subfigure}[t]{0.47\textwidth}
        \includegraphics[width=\textwidth]{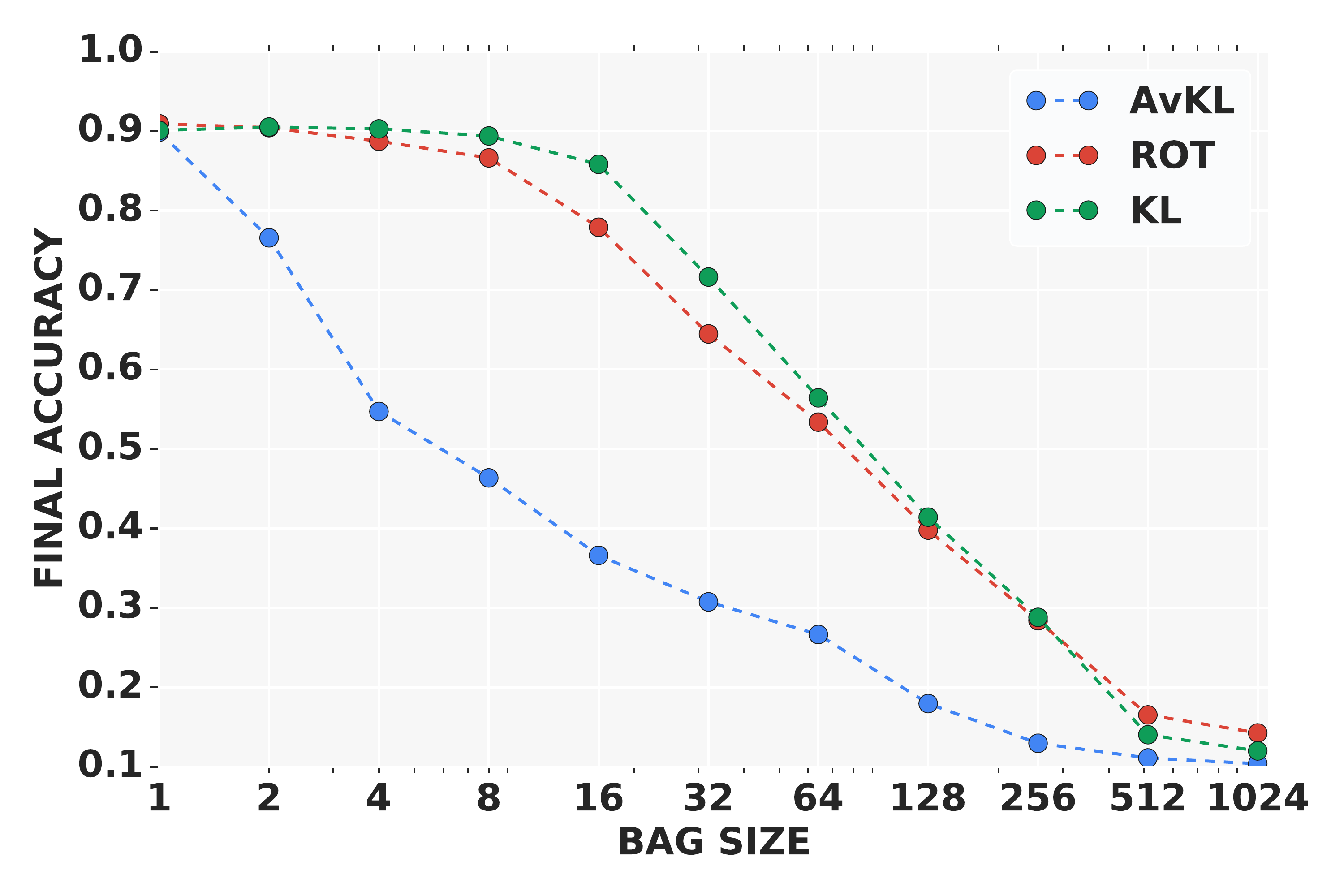}
        \caption{CIFAR 10}
        \label{fig:rot_kl_comp_10}
    \end{subfigure}
    ~ %
\begin{subfigure}[t]{0.47\textwidth}
        \includegraphics[width=\textwidth]{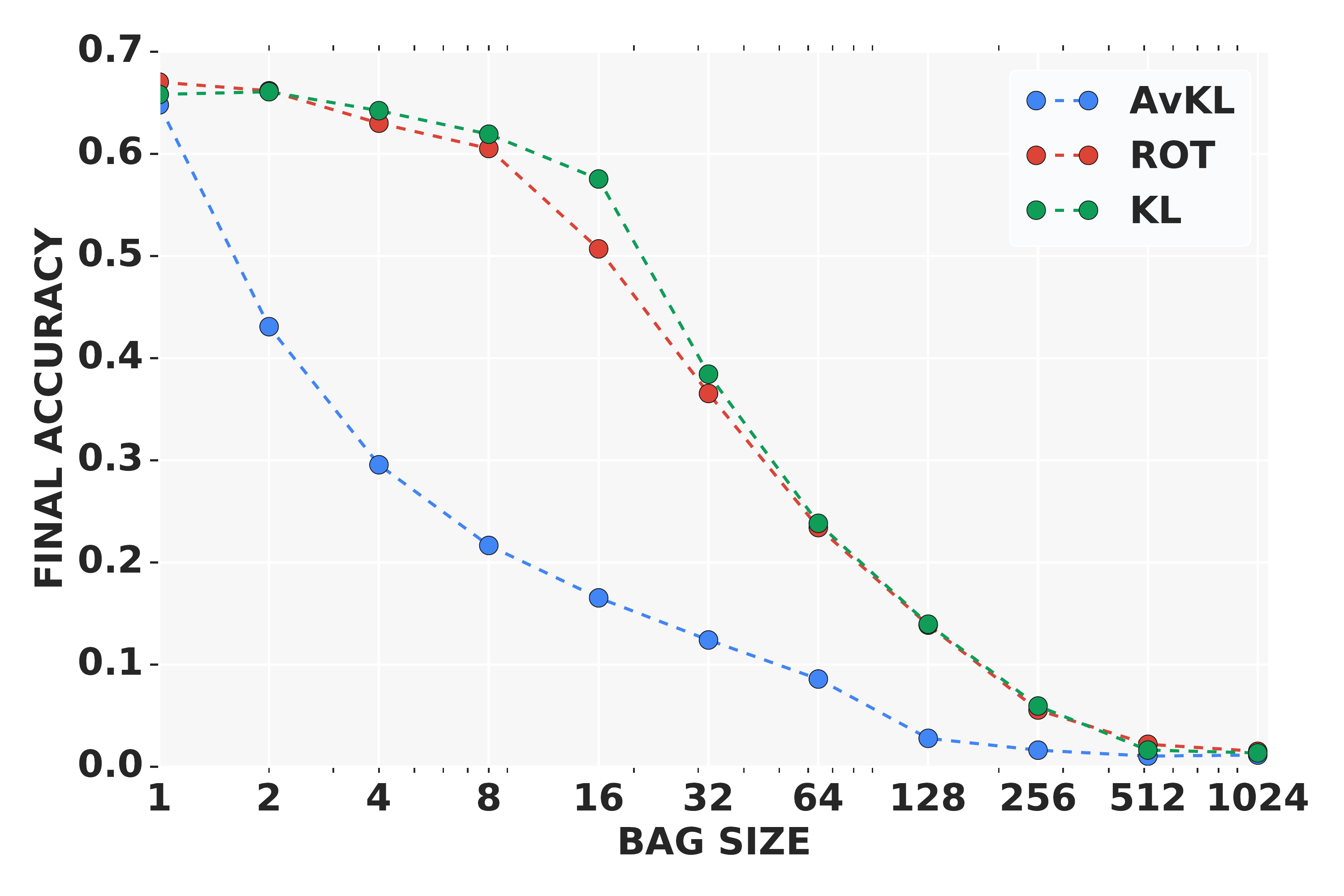}
        \caption{CIFAR 100}
        \label{fig:rot_kl_comp_100}
    \end{subfigure}
    \caption{Final test accuracy as a function of bag size for the baseline instance-level loss (AvgKL), and for the KL and ROT bag-level losses, on CIFAR10 (left) and CIFAR100 (right).}\label{fig:comp_kl_ot}
\end{figure}

{\bf Comparison of different losses} Figure \ref{fig:comp_kl_ot} compares the final accuracy of the different models for each bag size, on both CIFAR10 and CIFAR100 benchmarks. We first clearly see that the baseline method performs poorly, confirming the relevance of bag-level losses for LLP. Second, in both cases, we see that both bag-level losses perform very similarly overall, with a slight advantage for the KL loss over the ROT loss for a few bag sizes (4 to 64 on CIFAR10, 4 to 32 on CIFAR100), while the ROT loss slightly outperforms the KL loss for large bags ($512$ and $1024$) on CIFAR10. Overall, the fact that the simple KL loss tends to perform at least as well than the more evolved ROT loss suggests that jointly estimating  individual labels and the model parameters during training does not necessarily bring benefits over building a standard bag-level model. Finally, it is interesting to notice that in spite of the differences in difficulties between CIFAR10 and CIFAR100, the shape of the accuracy \emph{vs.} bag size curves is very similar between both benchmarks, with a sharp decrease in accuracy starting around 16 in bag size. This phenomenon suggests that the inflexion point of accuracy depending on the bag size is not that much task-dependent that it may be due to the model, its hyperparameters and its training scheme, as these are shared between our models trained on CIFAR10 and CIFAR100.

\section{Conclusion}
In this paper, we investigate the problem of learning from label proportions (LLP) in the barely explored setting of deep multi-class learning, most likely the current most active field of application in machine learning. We investigate two loss functions to address this problem: a modification of the standard cross-entropy, and a new loss function based on regularized optimal transport. On a proposed image classification task, we observe that our models are robust to switching from a fully supervised setting to the LLP one for bags of up to $16$ samples. For bigger bags, both loss functions suffer slow but steady degradations, and get close to the chance level for a bag size of a thousand samples. However, our new ROT loss based on optimal transport tends to show a higher robustness to big bags, which makes it more appropriate for real-world tasks which are likely to provide class statistics over large populations.

\section{Proofs}\label{sec:proof}
\subsection{Proof of Proposition~\ref{prop:fixedpoints}}
Our proof is based on standard arguments, following for example ~\citet{Cuturi2013Lightspeed,FrognerNIPS}, adapted to our setting. 
\begin{proof}
The solution $U^*$ of (\ref{eq:unbalancedLOSS}) is a saddle point of the Lagrangian
$$
L(U,\lambda) = \cbr{\frac{\alpha}{n} \left(\trace\br{C(\theta)^\top U} -\varepsilon H(U)\right)   + (1-\alpha) \KL\br{ U \One_{n}/n\,|\,z}} + \lambda^\top \br{U^\top \One_K - \One_n}\,,
$$
where $\lambda\in\RR^n$ is a vector of Lagrange multipliers for the constraint $U\in(\Delta_K)^n$. We derive, for any $(i,j)\in[1,K]\times[1,n]$:
$$
\frac{\partial L(U,\lambda)}{\partial U_{ij}} = \frac{\alpha}{n} C(\theta)_{ij} + \frac{\alpha\epsilon}{n}\log U_{ij} + \frac{1-\alpha}{n}\log\frac{[U\One_n]_i}{nz_i} + \lambda_j\,.
$$
Setting this derivative to $0$ for the saddle point $U^*$ gives:
$$
U_{ij}^* = e^{-\frac{C(\theta)_{ij}}{\epsilon}} \br{\frac{n z_i}{[U^* \One_n]_i}}^\frac{1-\alpha}{\alpha\epsilon} e^{-\frac{n\lambda_j}{\alpha\epsilon}}\,.
$$
Setting $K_{ij} = \exp\br{-\frac{C(\theta)_{ij}}{\epsilon}}$, $a = \br{{nz \oslash (U^*\One_n)}}^\beta$ with $\beta = \frac{1-\alpha}{\alpha\epsilon}$, and $b = \exp\br{-\frac{n\lambda}{\alpha\epsilon}}$, we finally get
$$
U_{ij}^* = K_{ij}a_i b_j\,,
$$
or in other words $U^* = \textrm{diag}(a) K \textrm{diag}(b)$. This implies that $U^* \One_n = a \otimes (Kb)$, therefore
$$
a = \br{nz \oslash \br{a \otimes (Kb)}}^\beta\,,
$$
from which we get
$$
a = \br{nz \oslash (Kb)}^{\tau}\,,
$$
with $\tau = \beta / (1+\beta) = \br{1+\frac{\alpha\epsilon}{1-\alpha}}^{-1}$.
Finally, since $(U^*)^\top \One_K = (K^\top a) \otimes b = \One_n$, we also get
$$
b = \One_n \oslash (K^\top a)\,.
$$
\end{proof}

\subsection{Proof of Proposition~\ref{prop:singleton}}
\begin{proof}
For a bag of size $1$, the label $z\in E_K$ has a unique non-zero coefficient, equal to $1$. Hence the KL term in (\ref{eq:unbalancedLOSS}) is infinite except when $U=z$, in which case the KL term is null and $H(U)=0$. We finally get that $\ell^{\text{relax-ent}}\br{F(\theta), z} = -\alpha \sum_{i=1}^K z_i \log f_\theta(x)_i $. 
\end{proof}

\bibliography{references}
\bibliographystyle{plainnat}

\end{document}